\documentclass[12pt,letterpaper]{article}
\usepackage[a4paper, total={7in, 10in}]{geometry}

\usepackage{graphicx}
\usepackage{helvet}
\usepackage{authblk}
\usepackage{hyperref}
\usepackage{amsmath} 
\usepackage{amssymb} 
\usepackage{orcidlink} 
\usepackage[super,comma,sort&compress]  
   {natbib}

\makeatletter
\renewcommand{\maketitle}{\bgroup\setlength{\parindent}{0pt}
\begin{flushleft}
  \textbf{\@title}
  
  \@author
\end{flushleft}\egroup}
\makeatother

\title{EL-PACA: Ensembled Low-dimensional Projections for Accurate Cell-type Annotation using deep learning for single-cell transcriptomics of human tissues and organoids}
\date{}

\author[1]{Muhammad Umar}
\author[2,3,*]{András Lakatos}
\author[1,*, ‡]{Arif Mahmood}
\author[2,*,‡]{Muhammad Asif}

\affil[1]{Information Technology University (ITU) of Punjab, Lahore, Pakistan}

\affil[2]{John van Geest Centre for Brain Repair, Department of Clinical Neurosciences, University of Cambridge, UK}
\affil[3]{MRC-WT Cambridge Stem Cell Institute, Cambridge Biomedical Campus, Cambridge, UK}

\affil[*]{Correspondence: András Lakatos (AL291@cam.ac.uk), Arif Mahmood (arif.mahmood@itu.edu.pk), Muhammad Asif (ma2129@cam.ac.uk).}
\affil[$\ddagger$]{Equal contributors}

\begin{document}

\maketitle

\section*{SUMMARY}

Single-cell transcriptomics has been revolutionising the discovery of human cell types and disease mechanisms. Accurate human cell-subtype annotation in complex neural tissues and disorders remains challenging due to the scarcity of postmortem tissue availability, in vitro culture-specific influences in surrogate models, and low cell abundance or transient states that arise during the disease process. Here, we present an Ensembled Low-dimensional Projections for Accurate Cell-type Annotation (EL-PACA), a computationally efficient framework that ensembles unsupervised principal component analysis with supervised multiple discriminant analysis to increase class separability prior to training a deep classifier. EL-PACA outperforms state-of-the-art methods on benchmarking datasets of blood and pancreatic cells. Importantly, it enables accurate, fine-grained cell-type identification in real-world datasets derived from human neural tissues. Applying EL-PACA to single-cell and single-nucleus RNA-sequencing datasets from postmortem human brain and 3D brain organoids, we show that combining supervised and unsupervised projections enables efficient, high-precision annotation that generalizes across tissues, technologies, and limited training data. Altogether,
EL-PACA offers an accessible alternative to computationally intensive foundation models, facilitating the discovery of rare or transient cell types relevant to complex disease mechanisms. The framework is openly available at https://github.com/umar1196/EL-PACA.

\section*{KEYWORDS}
single-cell RNA sequencing, cell type classification, machine learning, artificial intelligence (AI), data science, human tissues, brain organoids, neurodegeneration, amyotrophic lateral sclerosis (ALS)


\section*{INTRODUCTION}
Single-cell RNA-sequencing (scRNA-seq) has revolutionized biological research by enabling transcriptome profiling at cellular resolution \cite{r1}, thus allowing us to study complex processes in tissues. Specifically, it opened up opportunities for exploring mechanisms underlying cell differentiation during development \cite{r3} and cell state changes in pathology in several organs \cite{r5}. In particular, the application of scRNA-seq to postmortem brain or spinal cord tissues has advanced discoveries in human neuroscience, in which progress has been hindered due to the lack of suitable dynamic models or frameworks. While these advances have started to help identify disease-driving cells and pathological processes relevant to untreatable and fatal neurological conditions \cite{r6,r7}, cell type and cell state identification require further accuracy and precision, especially for transient or subtypes with fewer cells. 
A range of computational annotation approaches has emerged, spanning from simple marker gene-based and correlation-based methods \cite{r8,r9} to advanced dimensionality reduction techniques such as principal component analysis (PCA) \cite{r10}. More recently, deep learning and transfer learning approaches \cite{r11,r12}, as well as foundation models trained on millions of cells, have demonstrated significant promise in cell type classification \cite{r13,r15}.
Despite these advances, discovery through using scRNA-seq datasets for many neurological diseases, including amyotrophic lateral sclerosis (ALS), remains limited for several reasons. Firstly, scRNA-seq datasets of postmortem neural tissues from patients with ALS or other neurodegenerative conditions \cite{r6,r7,r23,r24,r25} are not yet robust. Thus, models trained on such data frequently suffer from domain mismatch and poor generalization due to data scarcity in light of human CNS sample availability issues. Secondly, transient cell types, as well as dynamically progressing pathological cell states that emerge during disease progression, often represent only a small fraction, making their identification challenging. Thirdly, emerging lab-grown neural tissue surrogates, such as human patient stem cell-derived brain organoids as disease models, offer an unprecedented source of data \cite{r16,r26}, but it is not without its caveats. While organoid systems provide valuable time-resolved single-cell transcriptomic data, the in vitro-in vivo transcriptional differences and the dynamic emergence of cell types across developmental timepoints introduce additional complexity for annotations.
To address these limitations, we introduce EL-PACA (Ensembled Low-dimensional Projections for Accurate Cell Type Annotation)—a computationally efficient, simple, and reference-informed method for accurate cell type classification in scRNA-seq data. We designed EL-PACA to leverage an ensemble of PCA and multiple discriminant analysis (MDA), so that it would efficiently capture sources of variance and maximize class separability in reduced-dimensional space. Unlike purely unsupervised methods, MDA incorporates known cell-type labels from reference datasets, thereby improving the discrimination of transcriptionally similar populations while training a deep neural network classifier for cell type classification.
We show that, by combining PCA and MDA projections with a streamlined deep neural network classifier, EL-PACA achieves high accuracy within a limited timescale without requiring substantial computational resources. We performed extensive benchmarking across gold-standard datasets derived from non-neural tissues or cells, independent queries, prior to analysing human postmortem brain tissue and in-house ALS brain organoid data.  We demonstrate that EL-PACA not only matches but often surpasses current state-of-the-art tools, particularly in distinguishing rare cell populations corresponding with subtypes or cell state changes. Thus, EL-PACA provides a novel tool for predicting both granular and major brain cell types with great precision without the requirement of large computing resources.

\section*{RESULTS}

\subsection*{EL-PACA, a two-step computational framework}
In order to improve classification in scRNA-seq datasets, we built a framework that captures both high variance in gene expression and maximizes cell-type or cell-state separation. To achieve this, we computed unsupervised and supervised lower-dimensional projections for scRNA expression matrix. Unsupervised projections were obtained by employing widely used principal component analysis (PCA), and for supervised projections, multiple discriminant analysis (MDA) was applied (Figure 1A).

Then, at the second step computed projections were used to train a deep neural network to predict the fine-grained clusters representing cell-types or states. Figure 1B illustrates the application of EL-PACA for cell type prediction in unseen (query) datasets. To annotate cell types in query data, EL-PACA first computes a lower-dimensional projection by mapping the query data onto the reference ensembled projections. In the second step, the query and reference projections are then integrated, and cell type labels are assigned to the query transcriptomic profiles based on the integrated representation (Figure 1B).

\subsection*{EL-PACA reveals the optimal deep neural network}
Designing a deep neural network for cell-type classification presents significant challenges related to underfitting and overfitting, potentially compromising the fidelity of the model. To address this, we systematically evaluated multiple classifier architectures using a published dataset deriving from Peripheral Blood Mononuclear Cells (PBMC1) representing ground truth. This included a varying number of hidden layers and nodes per layer (Table S1). Six distinct classifier architectures were assessed by comparing their performance on training and validation sets, with loss curves plotted for each configuration (Figure 2A, B). For each classifier, 20\% of the training data was held out as a validation set. All classifier architectures were trained for 100 epochs, using a combined input of 100 principal components (PCs) and 9 MDA components (matching the number of unique cell types in the dataset). To further assess the fidelity of our model, F1 scores were calculated on the test data. The F1 score represents both precision (the proportion of correctly predicted cell types among all predictions) and recall (the proportion of actual cell types that were correctly identified). The F1 score is a more useful metric for cell type classification, as it provides a balanced assessment when there is an imbalance in the proportion of cell types (Table S1).
Classifiers 1 and 2, each with two hidden layers and lower node counts (64–16 and 64–32–16, respectively), showed lower performance (Figure 2B). Classifiers 3 and 4, which had higher node counts, achieved improved F1 scores, albeit classifier 4 yielded the lowest F1 score among the high-capacity models despite its complexity (Table S1). Classifiers 5 and 6 were designed to resemble autoencoders, with classifier 6 featuring a greater number of nodes in each hidden layer. Notably, classifier 6 achieved the highest F1 score, indicating superior performance in minimizing both false positives and false negatives (Figure 2B). Although classifier 3 had the lowest validation loss, its simpler architecture resulted in a lower F1 score compared to classifier 6. Classifier 6 was used for all subsequent analyses presented in this manuscript.


\subsection*{EL-PACA enables robust cell type classification using ensembled PCA and MDA}
To investigate the impact of dimensionality reduction on cell type classification performance, we computed low-dimensional representations using both PCA and MDA. For MDA, the number of components was set equal to the number of unique cell types in the training set. We systematically varied the number of principal components (PCs) selected from PCA, combining these with the MDA components to evaluate the effect on classification accuracy and F1 score. (Figure 2C, D).
To comprehensively evaluate the robustness of EL-PACA for varied number of PCs, we designed two experimental scenarios. First, we assessed performance using a subset of the training data with known labels—a setup referred to as reference-to-reference prediction. Second, we trained the classifier on the complete reference data and tested it on entirely unseen query datasets, a scenario termed reference-to-query prediction. Prior to classifier training, all datasets underwent standard preprocessing and normalization, with the 2,000 most highly variable genes selected for downstream analysis. Table S2 provides the details on ground truth datasets.  
For PBMC1 cell type classification, we observed a marked improvement in F1 score when PCA components were combined with MDA components (Figure 2C). Using only MDA components, the F1 score was 69.01\%; this increased substantially to a maximum of 95.97\% upon ensembling with 70 PCs, highlighting the benefit of integrating PCA-derived variation for optimal classification. In the Baron dataset, the optimal ensemble included 20 PCs with MDA components to achieve the highest F1 score of 98.06\% compared to when PCs were 10 or less. For the integrated datasets (Muraro, Xin, and Baron, integrated using the Harmony method \cite{r31}), combining 100 PCs with MDA components yielded a maximum F1 score of 98.64\% (Figure 2C). These results demonstrate that integrating both PCA and MDA components significantly improves cell type classification accuracy for reference-to-reference prediction, compared to using MDA or PCA components alone.
In the reference-to-query scenario, query datasets were pre-processed as described (Figure 1B). When predicting PBMC2 cell types using a model trained on PBMC1, EL-PACA achieved an F1 score of 92.85\%, utilizing 90 PCA components in combination with MDA.
When the Baron reference served as the query dataset (with the remaining three datasets as reference), EL-PACA achieved an F1 score of 97.22\%. Similarly, for the Muraro dataset as the query, ensembling 90 PCA components with MDA further improved performance, as shown in Figure 2D. For the Segerstolpe dataset, adding PCA components provided comparable performance to using MDA alone, suggesting that the benefit of PCA-MDA ensembling may vary with dataset complexity and cell type composition. Overall, these analyses demonstrate that ensembling lower-dimensional projections from both MDA and PCA consistently improves cell type classification across a range of datasets and prediction scenarios.

\subsection*{EL-PACA outperforms existing cell type classification methods}
We benchmarked EL-PACA across six diverse single-cell RNA-seq datasets that varied in sequencing technologies, sample sizes, and known cell type complexity (Table S2). To contextualize the performance of EL-PACA, we compared it with several widely adopted cell type annotation methods: ACTINN \cite{r12}, CaSTLe \cite{r11}, SingleR \cite{r9}, and scPred \cite{r10}. These methods span a range of analytical paradigms, including multi-layer neural networks (ACTINN), correlation-based annotation (SingleR), SVD and SVM-based classification (scPred), and transfer learning (CaSTLe). Therefore, the diversity of these approaches could provide a robust benchmark for evaluating the strengths of EL-PACA across different data analysis strategies.
We first assessed the performance of each method in classifying held-out subsets from three reference datasets (PBMC1, Baron, and a combined set of Muraro, Xin, and Baron datasets), using 80\% of the data for training and 20\% for testing. Across all reference prediction experiments, EL-PACA consistently outperformed or matched state-of-the-art methods (Figure 2E). For PBMC1, EL-PACA achieved the highest F1 score (95.97\%), followed by SingleR (92.27\%). While other methods did not perform well. In the Baron dataset, EL-PACA again demonstrated superior performance (F1 score: 98.06\%), with SingleR (96.54) and CaSTLe (72.89) ranking second and third, respectively. On the combined Muraro, Xin, and Baron dataset, the performance of EL-PACA, SingleR, and CaSTLe was comparable, while ACTINN and scPred were inferior (Figure 2E).
We next evaluated the reliability of the model by testing prediction performance on unseen query datasets, simulating real-world scenarios where new data may differ substantially from training references. For these experiments, lower-dimensional projections were computed for reference data separately, followed by the projection of query data into lower dimensions using reference projections. The Harmony method was applied to minimize batch effects between reference and query profiles. EL-PACA was trained on reference projections, while the query dataset was used solely for prediction (Figure 2F).
First, four human datasets from pancreatic tissue samples (Muraro, Xin, Segerstolpe, and Baron) were used to test cross-dataset prediction. In each experiment, one dataset (e.g., Muraro) served as the query, while the remaining datasets (e.g., Xin, Segerstolpe, and Baron) formed the reference. EL-PACA, CaSTLe, and ACTINN showed comparable performance when Muraro was used as the query, outperforming SingleR (92.04\%) and scPred (8.44\%). When Baron was the query, EL-PACA, SingleR, and scPred delivered the best results compared to CaSTLe, and ACTINN. In the third experiment, where Segerstolpe was the query, EL-PACA remained the top performer, with SingleR and scPred ranking second and third, respectively. To assess performance in blood cell-type classification, PBMC1 was used as the reference and PBMC2 as the query, with nine shared cell types. EL-PACA achieved an F1 score of 92.85\%. While SingleR and scPred produced a slightly lower F1 score, 87.31\% and 84.99, respectively. This was notably higher than CaSTLe, which was ranked lowest in both F1 score (Figure 2F). Overall, these results demonstrate that EL-PACA not only matches but frequently exceeds the accuracy of current state-of-the-art methods in both within-dataset and cross-dataset scenarios (Figure 2E, F).

\subsection*{EL-PACA accurately predicts major cell types in patient-derived postmortem brain samples}

Following the benchmarking of EL-PACA in standard non-neural tissue datasets, we then applied our framework to neural tissue transcriptomes, which are increasingly used for aiding mechanistic discovery of neurological diseases. For this, the use of postmortem brain samples remains challenging, given that the commonly applied single-nucleus (sn)RNA-seq studies do not capture the whole transcriptome. This issue may compromise results, especially in the context of rare cell types and subtle disease-related cell state changes. In this study, we performed a comparative evaluation between EL-PACA and two other top-ranking cell-type classification methods (CaSTLe and SingleR) using a postmortem control and ALS brain snRNA-seq dataset \cite{r7}. The three methods were trained on snRNA-seq profiles to predict fourteen major cell-types (Figure 3A). 
In predicting the major cell types, EL-PACA consistently outperformed both SingleR and CaSTLe, achieving the highest mean F1 score across all folds (Figure 3A, B). CaSTLe ranked second but showed greater variation in performance. SingleR demonstrated lower overall performance and the highest variability in F1 score across folds. These results highlight the superior performance and robustness of EL-PACA in classifying major brain cell types even when cell type even when cell type-specific transcriptomes are influenced by neurodegenerative pathologies (Figure 3B).
For major cell types, including astrocytes, OPCs, microglia, and oligodendrocytes, all the methods showed comparable performance. Overall, for these cells, the F1 score of each method was higher than 95\%. A similar pattern in performance was also observed for excitatory neurons, such as deep and intermediate excitatory neurons (Figure 3B). Altogether, in datasets including a large number of cells, the performance of EL-PACA, SingleR, and CaSTLe was comparable in predicting major cell-type populations in ALS postmortem brain samples.

Notably, EL-PACA demonstrated higher precision across all cell types, even those with low cell abundance (Figure 3C, D). For rare and underrepresented cell types with fewer than 100 cells—such as “other CGE-derived inhibitory neurons”, vascular and leptomeningeal cells—EL-PACA achieved slightly better classification than both CaSTLe and SingleR. A similar performance trend was observed for other cell types with fewer than 2,000 cells, though modest cross-assignment was observed among closely related subtypes across all methods. (Figure 3B, D).

\subsection*{EL-PACA accurately predicts fine-grained cell types with low abundance in the postmortem brain}
We evaluated EL-PACA, SingleR, and CaSTLe on 49 cellular subtypes annotated by Li et al. \cite{r7} spanning a wide range of abundances. Li et al. performed single-nucleus RNA sequencing (snRNA-seq) on postmortem brain samples from control individuals and patients with ALS, followed by a deeper annotation of cellular states using specific markers.   
Rare subpopulations are challenging to annotate owing to transcriptional overlap with larger populations. All three methods were trained on the same ALS postmortem brain dataset re-annotated at the subtype level. Low-abundance subtypes included excitatory and inhibitory neuronal subpopulations, as well as non-neuronal types (e.g., oligodendrocytes and astroglia), with substantial cell number variations (Supplementary Figure. 1A, 2A, 3A). As in previous analyses, we used stratified five-fold cross-validation and the F1 score to assess performance across all subtype classes.
EL-PACA showed consistent performance across folds, albeit with a wider F1 score range than in the major cell-type task, reflecting the increased difficulty of subtype resolution. SingleR was stable but consistently below EL-PACA, whereas CaSTLe performed the lowest overall. Mean F1 score across five folds was 0.93 for EL-PACA, compared with 0.86 for SingleR and 0.75 for CaSTLe (Figure 4A).
To verify performance on the least-abundant populations, we summarised per-method performance for the ten least-abundant excitatory subtypes (Figure 4B), the ten least-abundant inhibitory subtypes (Figure 4C), and non-neuronal subtypes (Figure 4D), with examples of per-subtype confusion patterns and misclassifications (Supplementary Figures 1–3). While EL-PACA and SingleR showed broadly comparable accuracy for very small subtypes ($\sim\leq200$ cells), the former was superior when subtype sizes exceeded $\sim\leq200$ cells, whereas, in general, CaSTLe showed an inferior performance. A similar trend was held for inhibitory neuronal subtypes, which ranged from a few cells to $\sim\leq250$ cells. For non-neuronal cells, which were generally more abundant than neurons, the performance of EL-PACA was comparable to or slightly better than that seen for SingleR and CaSTLe. For example, EL-PACA achieved an F1 of 93.9\% on CD44-expressing astrocytes, versus 71.9\% for SingleR and 81.2\% for CaSTLe.
Across subpopulations representing cell subtypes or states, the above metrics indicate that EL-PACA more reliably captures discriminative transcriptional signatures in scarce populations than other methods used for comparison (Supplementary Figures. 1–3).

\subsection*{EL-PACA accurately captures cell-type diversity in brain organoid tissue, despite culture conditions and disease-related perturbations}

Next, we evaluated the performance of EL-PACA using datasets derived from human in vitro 3D brain organoids grown from control and ALS patient-derived stem cells. While this is a promising surrogate model for disease-related discoveries, accurate cell type annotations can be challenging due to the in vitro-in vivo transcriptional dichotomy as well as pathologic influences on the transcriptome. To do so, we compared EL-PACA with other methods in their ability to capture the fine-grained subsets of neuronal and glial cell populations. For this, initially we trained EL-PACA, along with SingleR and CaSTLe, and utilised five-fold stratified cross-validation on our previously published ALS brain organoid scRNA-seq data \cite{r16} using F1 score as a metric. Overall, EL-PACA achieved the highest mean F1 score for cell-type classification from ALS brain organoids, outperforming both SingleR and CaSTLe (Figure 5A-C).
Specifically, EL-PACA achieved over 90\% accuracy in identifying key cortical cell types, including deep-layer (5/6) and upper-layer (2/3 and 3/4) excitatory neurons, interneurons, and oligodendrocyte/OPC populations (Figure 5B). Figure 5B also displays the predicted classes for misclassified cells—those whose predicted cell type does not match the ground truth—showing which cell types they were incorrectly assigned to. The misclassified deep-layer (5/6) excitatory neurons were primarily assigned to the upper layer (2/3 and 3/4) or immature excitatory neurons. Similarly, upper-layer excitatory neurons were sometimes misclassified as other upper-layer types, deep-layer neurons, or immature excitatory neurons. The performance of SingleR was lower than that of EL-PACA in identifying deep layer (87\%) and upper layer (2/3) excitatory neurons (71\%), upper layer (3/4) excitatory neurons (96\%), as well as interneurons (~89\%). CaSTLe performance on classifying the deep layer neurons was lowest compared to EL-PACA and SingleR.   
We then tested the performance of EL-PACA on certain intermediate or transitional cell states with a small number of cells (Figure 5D). Notably, intermediate radial glia (iRG) and truncated radial glia (tRG) cells were accurately classified at 97\% and 89\%, respectively. This outperformed SingleR (iRG: 61\%, tRG: 75\%) and CaSTLe (iRG: 90\%, tRG: 81\%). However, EL-PACA underperformed for outer radial glia (oRG) (60\%) compared to SingleR (84\%) but still did better than CaSTLe (38\%). These misclassifications were largely confined to developmentally related or transcriptionally similar populations. For instance, oRG and tRG are both radial glial subtypes involved in neurogenesis and gliogenesis, and their gene expression overlaps with early gliogenic markers seen in astrocytes and OPCs. This likely contributes to prediction ambiguity. The oRG cell cluster contains 650 cells, suggesting that EL-PACA’s performance may be influenced by cluster size. Notably, the choroid plexus cluster, which has the fewest cells in the dataset (n = 264), was still accurately identified by EL-PACA, achieving an F1 score of 96\% despite the limited cell number (Figure 5D).

Furthermore, EL-PACA outperformed other methods in identifying astroglial, oligodendrocyte/OPC, interneurons, and intermediate progenitor cell populations (Figure 5B, D). While EL-PACA demonstrates high precision in classifying mature neuronal and glial populations, choroid plexus cells were also effectively detected. A few of these cells were misclassified as astroglia, tRG, or oRG, which may be biologically plausible due to the developmental similarities.
Overall, EL-PACA demonstrated strong and consistent performance in predicting cell types in brain organoids, showcasing its potential as a versatile tool for accurate cell type annotation across diverse brain-derived datasets, despite the in vitro culture environment.

\section*{DISCUSSION}

Single-cell RNA sequencing (scRNA-seq) has revolutionized biological discovery by enabling transcriptomic profiling of millions of cells at unprecedented scale and resolution \cite{r1}. It has become a widely used and promising tool for studying human biology and pathobiology in complex tissues and diseases, with emerging challenges in cell type annotation. Here, we introduce EL-PACA, a simple knowledge-integrated framework for cell type prediction. Notably, its design is compatible with widely used pipelines such as Seurat \cite{r17} and Scanpy \cite{r18}, ensuring seamless integration into existing workflows. We show that, unlike many existing approaches, EL-PACA is robust, yet easily applied, requiring neither extensive computational resources nor advanced programming expertise. We demonstrate that EL-PACA possesses a particularly superior performance in identifying rare cell types or states, despite neurodegenerative cues. Overall, our approach provides a practical and accurate solution for the scientific community, especially for those with limited computing resources that restrict the use of deep foundation models.

Mainly, EL-PACA consists of two components: (i) dimensionality reduction of cellular expression profiles using an ensemble of PCA and MDA projections, and (ii) classifier training for predicting cell types in both reference data and unseen query datasets. Ground-truth datasets were used to extensively benchmark these components. The results demonstrated that EL-PACA effectively captures biological variation through its ensembled projections, which in turn enable the development of a compact yet accurate neural network classifier for cell type prediction. 

Following a systematic benchmarking using gold-standard reference datasets of blood and pancreatic tissue cells, EL-PACA demonstrated superior performance on real-world applications, including postmortem brain and also organoid tissues affected by complex disease phenotypes \cite{r6,r7,r16}. This is of particular significance as the scarcity of high-quality datasets and the diverse dynamically changing cell states makes cell classification particularly challenging. Existing tools, often optimized for healthy tissues, frequently underperform when applied to disease-compromised samples \cite{r27,r32}. In contrast, EL-PACA matched or outperformed state-of-the-art methods across both neuronal and non-neuronal populations, including rare and fine-grained subtypes that represent only a small fraction of total cells \cite{r6,r7,r28,r29}. These populations, further confounded by pathological processes, are difficult to resolve, but EL-PACA consistently captured them more effectively than alternative methods. Occasional subtle misclassifications likely reflect biological ambiguity rather than methodological limitations. By incorporating cell type labels during training, EL-PACA enhances class separability and enables detection of biologically meaningful sub-signatures that may underlie disease mechanisms.

We further validated EL-PACA using dataset of brain organoids derived from ALS patient specific iPSC lines. Brain organoids provide experimentally tractable in vitro models that capture both human cortical cell type diversity and disease-related molecular hallmarks \cite{r16, r33, r34, r35, r36}. Thus, cortical organoids offer a unique opportunity to study disease-related processes. However, distinguishing transcriptionally similar subpopulations, such as deep- and upper-layer excitatory neurons, especially when influenced by disease states, remains challenging. In this context, EL-PACA maintained greater biological coherence in its predictions and showed improved resolution of cortical populations. This applied computational approach helps improve the fidelity of brain organoids for identifying and studying developmental and disease phenotypes.
Beyond its strong biological applications, EL-PACA is computationally efficient, requiring minimal programming expertise or high-performance computing resources. In contrast, correlation-based annotation methods often lack scalability, while unsupervised machine learning approaches (e.g., PCA-based classifiers) frequently struggle with fine-grained cell type resolution \cite{r27}. Although large-scale foundation models offer impressive generalization \cite{r13,r14,r15}, their high computational demands and need for fine-tuning limit accessibility, particularly in disease-specific or temporally resolved datasets. EL-PACA addresses these challenges by explicitly integrating known cell type information during training, thereby enhancing class separability and improving classification accuracy. Specifically, by combining supervised MDA with unsupervised PCA projections, EL-PACA leverages both prior biological knowledge and data-driven structure to train a deep neural network classifier.
Despite these significant advantages, EL-PACA also has limitations. As with other supervised approaches, its performance depends on the quality and diversity of the reference dataset. For ALS, where annotated resources remain limited, broader adoption will require expanded high-resolution datasets. Additionally, while EL-PACA currently focuses on expression-based classification, future iterations could integrate complementary biological knowledge, such as protein–protein interaction networks, gene regulatory maps, or multi-omic data, to further improve interpretability and accuracy. Combining the knowledge-guided framework of EL-PACA with the generalization capacity of foundation models may ultimately enable deeper insights into disease progression and subtype-specific mechanisms.
In conclusion, EL-PACA provides a robust and resource-efficient solution for cell type classification in single-cell transcriptomics, especially for complex human neural tissues. Its strong performance across benchmarking datasets and real-world data, together with ease of integration into existing workflows, establishes EL-PACA as a valuable and widely applicable tool for accurately identifying cell types influenced either by the culture environment or disease states.

\newpage

\section*{METHODS}

Figure 1 provides an overview of the proposed EL-PACA framework. The method consists of (i) preprocessing and normalization of single-cell expression data, (ii) dimensionality reduction using a combined projection of principal component analysis (PCA) and multiple discriminant analysis (MDA), (iii) training a compact deep neural network on the ensembled features, and (iv) predicting cell labels for unseen data (Figure 1B).

\subsection*{Data prepossessing and normalization}

Single-cell gene expression matrices were filtered and preprocessed using the \texttt{Scanpy} \cite{r18}. Genes with highly variable expression across cells were selected by computing mean and variance, and genes with a high variance-to-mean ratio were retained. Expression values were log-normalized and scaled prior to downstream analysis.

\subsection*{Dimensionality reduction}

The dimensionality of the pre-processed gene expression matrix $X_{train}$ was reduced using both multiple discriminant analysis (MDA) and principal component analysis (PCA) (Figure 1A). The projections obtained from the two methods were combined to form an ensemble representation, which was then used to project the data into a lower-dimensional space. The model was trained on this ensemble projection of the training set, and performance was evaluated on a separate validation set.

For PCA, the pre-processed gene expression matrix $X_{train}$, consisting of $n$ cells and $m$ genes, was centered by subtracting the mean expression value $\mu$ of each gene from its corresponding values (equation 1). Principal components capturing the highest variance were then extracted for downstream analysis.

\begin{equation}
    \mu = \frac{1}{n} \sum \limits_{i=1}^{n} x_{i}
\end{equation}

\[X_{nm} = \begin{bmatrix} 
    x_{11}-\mu_{1} & x_{12}-\mu_{2} & \dots & x_{1m}-\mu_{m}\\
    x_{21}-\mu_{1} & x_{22}-\mu_{2} & \dots & x_{2m}-\mu_{m}\\
    \vdots & \vdots &\ddots &\vdots & \\
    x_{n1}-\mu_{1} & x_{n2}-\mu_{2} & \dots & x_{nm}-\mu_{m} 
    \end{bmatrix}
\]

The covariance matrix $C$ was then calculated, measuring how changes in one variable relate to changes in another (equation 2).

\begin{equation}
    C_{mm} = X_{nm}^{\top} * X_{nm}
\end{equation}

\[C_{mm} = \begin{bmatrix} 
    C(x_{11},x_{11}) & C(x_{12},x_{22}) & \dots & C(x_{1m},x_{mm})\\
    C(x_{21},x_{11}) & C(x_{22},x_{22}) & \dots & C(x_{2m},x_{mm})\\
    \vdots & \vdots &\ddots &\vdots & \\
    C(x_{m1},x_{11}) & C(x_{m2},x_{22}) & \dots & C(x_{mm},x_{mm}) 
    \end{bmatrix}
\]

The eigenvalue equation was solved to obtain the eigenvectors of the covariance matrix. First, eigenvalues were computed, and the eigenvectors corresponding to the largest eigenvalues were identified (equation 3):

\begin{equation}
    CP = \lambda P
\end{equation}

where $C$ is the covariance matrix, $P$ is the eigenvector, and $\lambda$ is the eigenvalue corresponding to eigenvector $P$.

The eigenvectors were sorted in descending order, and those associated with the largest eigenvalues were selected, as they capture the highest variance in the data. The matrix $U$, containing the $l$ components with high variance from the matrix $P$ for $X_{train}$, is represented as:

\[U_{ml} = \begin{bmatrix} 
    u_{11} & u_{12} & \dots & u_{1l}\\
    u_{21} & u_{22} & \dots & u_{2l}\\
    \vdots & \vdots &\ddots &\vdots & \\
    u_{m1} & u_{m2} & \dots & u_{ml} 
    \end{bmatrix}
\]

Multiple discriminant analysis (MDA) was used to minimize within-class scatter and maximize between-class separation, thereby improving class separability in the lower-dimensional space. To calculate the between-class scatter $S_{b}$, the mean of each class was computed, subtracted from the overall mean $m$, multiplied by its transpose, and scaled by the number of samples $n_i$ in class $i$. Summing across classes yielded $S_{b}$ (equation 4):

\begin{equation}
    S_{b} = \sum_{i=1}^{k} n_i (m_{i}-m)(m_{i}-m)^{\top}
\end{equation}
where $k$ is the number of classes and $m_{i}$ is the mean of class $i$.

Within-class scatter was computed to represent the variance of data points around their respective class means. For each class, the mean vector was subtracted from the class data, and the result was multiplied by its transpose to obtain $S_i$. The sum of these matrices produced the overall within-class scatter matrix $S_{w}$ (equations 5–6):

\begin{equation}
    S_{w} = \sum _{i=1}^{k}S_{i} 
\end{equation}
Where $S_{i}$ is
\begin{equation}
    S_{i} = \sum_{x\in D_{i}} (x-m_{i})(x-m_{i})^{\top}
\end{equation}

Here $x$ and $m_{i}$ are the data points and class mean for the $i^{th}$ class, respectively, and $S_{w}$ is the within-class scatter matrix.

After obtaining both scatter matrices, the generalized eigenvalue problem was solved to compute the discriminant components (equation 7):

\begin{equation}
    S_{w}^{-} * S_{b} * V = \lambda * V
\end{equation}

where $S_{w}^{-1}$ is the inverse of the within-class scatter matrix, $\lambda$ are the eigenvalues, and $V$ is the matrix of eigenvectors. The eigenvectors were sorted in descending order, and the number of retained components was set equal to the number of unique classes in the dataset. For a dataset with $k$ unique classes and $n$ records, the discriminant components are represented as:

\[V_{mk} = \begin{bmatrix} 
    v_{11} & v_{12} & \dots & v_{1k}\\
    v_{21} & v_{22} & \dots & v_{2k}\\
    \vdots & \vdots &\ddots &\vdots & \\
    v_{m1} & v_{m2} & \dots & v_{mk} 
    \end{bmatrix}
\]

\subsection*{Deep classifier training}

After obtaining both the discriminant components and the high-variance components, they were combined to form an ensemble representation $S$:

\[S_{mj} = \begin{bmatrix} 
    V \space U
    \end{bmatrix}
\]

or equivalently,

\[S_{mj} = \begin{bmatrix} 
    v_{11} & v_{12} & \dots & v_{1k}&u_{11} & u_{12} & \dots & u_{1l}\\
    v_{21} & v_{22} & \dots & v_{2k}&u_{21} & u_{22} & \dots & u_{2l}\\
    \vdots & \vdots &\ddots &\vdots & \vdots & \vdots &\ddots &\vdots & \\
    v_{m1} & v_{m2} & \dots & v_{mk}& u_{m1} & u_{m2} & \dots & u_{ml}
    \end{bmatrix}
\]

Here, $S$ is an $n \times j$ matrix, where $n$ is the number of rows and $j$ is the sum of $k$ and $l$. This ensemble representation was then used to project $X_{nm}$ into a lower-dimensional space:

\begin{equation}
    P_{nj} = X_{nm} * S_{mj}
\end{equation}

The lower-dimensional representation $P$ was subsequently used to train a feedforward neural network consisting of multiple hidden layers and an output layer. The output of the hidden layers $H$ was defined as:

\begin{equation}
    H = \sigma(W^{5} * \sigma(W^{4} * \sigma(W^{3} * \sigma(W^{2} * \sigma(W^{1} * P + b^{1}) + b^{2}) + b^{3}) + b^{4}) + b^{5})
\end{equation}

where $P$ is the input matrix, $W^{1}, W^{2}, W^{3}, W^{4}, W^{5}$ and $b^{1}, b^{2}, b^{3}, b^{4}, b^{5}$ are the weights and biases of the five hidden layers, and $\sigma(.) = \mathrm{ReLU}(.)$ is the non-linear activation function.

The classifier output was defined as:

\begin{equation}
    \hat{Y} = softmax(HW^{(o)}+b^{(o)})
\end{equation}

where $W^{(o)}$ and $b^{(o)}$ are the weights and biases of the output layer, $softmax$ is the activation function, and $\hat{Y}$ represents the predicted class probabilities.

Classifier training aimed to minimize the cross-entropy loss between the ground-truth and predicted labels. The loss function was defined as:

\begin{equation}
    J = - \frac{1}{N} (\sum _{i=1} ^{N} \sum_{j=1}^{J} y_{ij}. \log (\hat{y_{ij}}))
\end{equation}

where $N$ is the total number of training examples, $y_{ij}$ represents the true label, and $\hat{y_{ij}}$ represents the predicted probability that the $i^{th}$ record belongs to class $j$.

\subsection*{Cell type prediction}

The gene expression matrix of the test data, consisting of $q$ records and 2000 variable genes, is represented as $T_{qm}$ and projected into the lower-dimensional space as:

\begin{equation}
    \hat{P}_{qj} = T_{qm} * S_{mj}
\end{equation}

Here, $\hat{P}_{qj}$ is the lower-dimensional representation of the test data. The trained neural network then used this representation to predict the cell type labels.


\newpage


\section*{RESOURCE AVAILABILITY}


\subsection*{Lead contact}


Requests for further information and resources should be directed to and will be fulfilled by the lead contact, Muhammad Asif: ma2129@cam.ac.uk.

\subsection*{Materials availability}


Not applicable. 
\subsection*{Data and code availability}

\begin{itemize}
    \item This study did not generate any new sequencing datasets.
    \item The datasets used to evaluate the performance of EL-PACA are publicly available in the Gene Expression Omnibus (GEO) under accession numbers GSE219281 and GSE180122.
    \item The EL-PACA code is openly available at: https://github.com/umar1196/EL-PACA.    
\end{itemize}

\section*{ACKNOWLEDGMENTS}

The project was supported by the UKRI Medical Research Council (MRC) grant awarded to AL (MR/X006867/1).
\section*{AUTHOR CONTRIBUTIONS}

MA and AM conceived the study. MU developed the methodology and performed the analysis. All authors reviewed and interpreted the results. The first draft of the manuscript was written by MU and MA. AL wrote parts and provided critical revisions. MA and AM supervised the project, with input from AL on the application of EL-PACA to human neural tissue biology.

\section*{DECLARATION OF INTERESTS}


Not applicable

\section*{DECLARATION OF GENERATIVE AI AND AI-ASSISTED TECHNOLOGIES}


Not applicable

\section*{SUPPLEMENTAL INFORMATION INDEX}




\begin{description}
  \item Table S1. EL-PACA architecture. F1 score on test set of PBMC1 data using different neural network architecture.
  \item Table S2. Ground truth datasets for performance evaluation
  \item Figures S1-S3 
\end{description}

\newpage

\section*{MAIN FIGURE TITLES AND LEGENDS}




\noindent\includegraphics[width=1\linewidth]{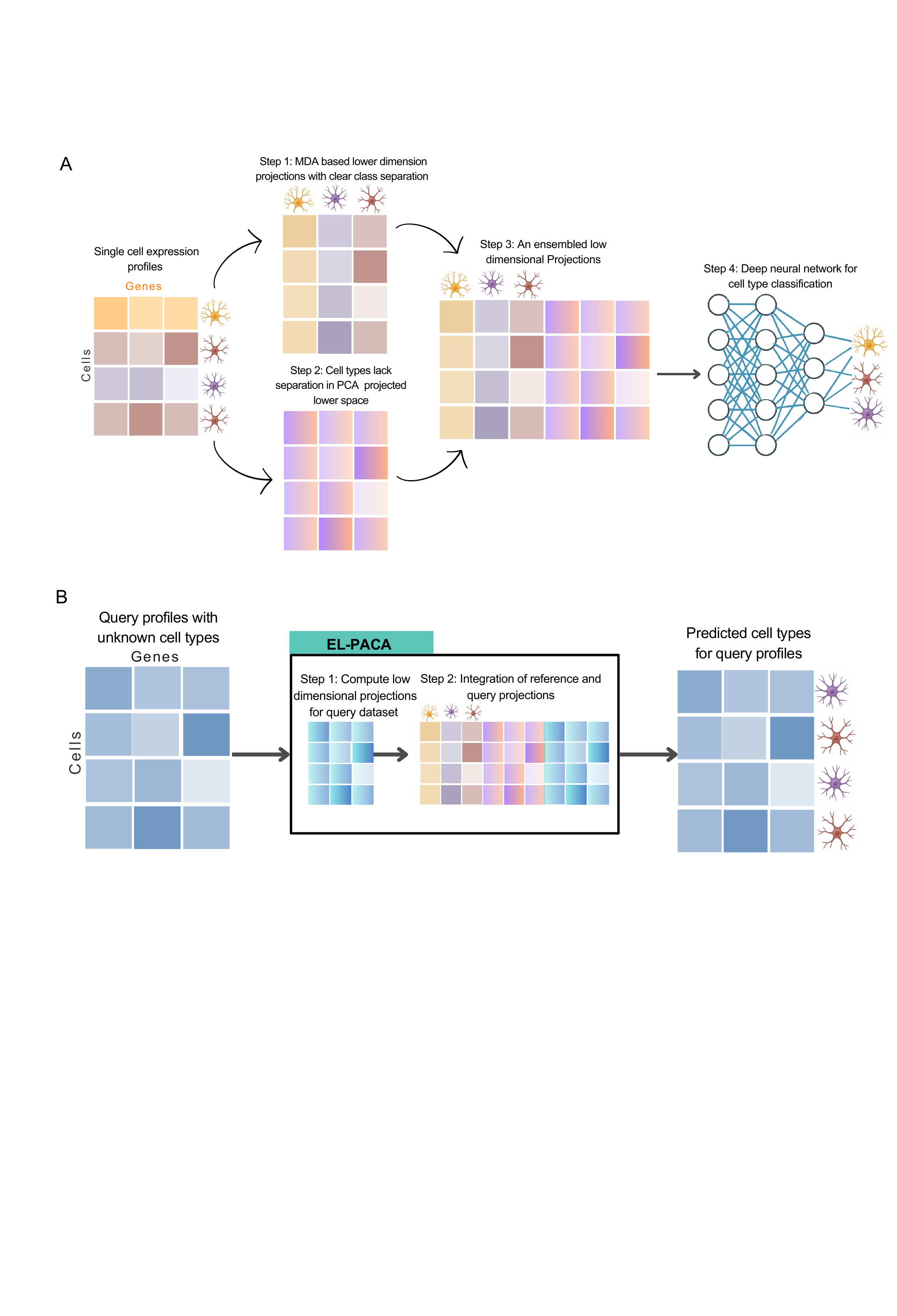}

\subsection*{Figure 1. Schematic overview of the EL-PACA architecture.}

\textbf{A} Training phase. EL-PACA builds a low-dimensional representation by ensembling unsupervised (PCA) and supervised (MDA) projections learned from a reference dataset. The concatenated projections are input to a deep neural network classifier. Performance is evaluated with stratified five-fold cross-validation: in each fold, the model is trained on four folds and tested on the held-out fold; the F1 score is used as the primary metric. \textbf{B} Inference on a query dataset. EL-PACA identifies the intersection of highly variable genes between reference and query, projects the query into the reference PCA/MDA spaces, integrates reference and query with Harmony \cite{r31} to correct batch effects, and assigns cell-type labels using the trained classifier.

\newpage

\noindent\includegraphics[width=1\linewidth]{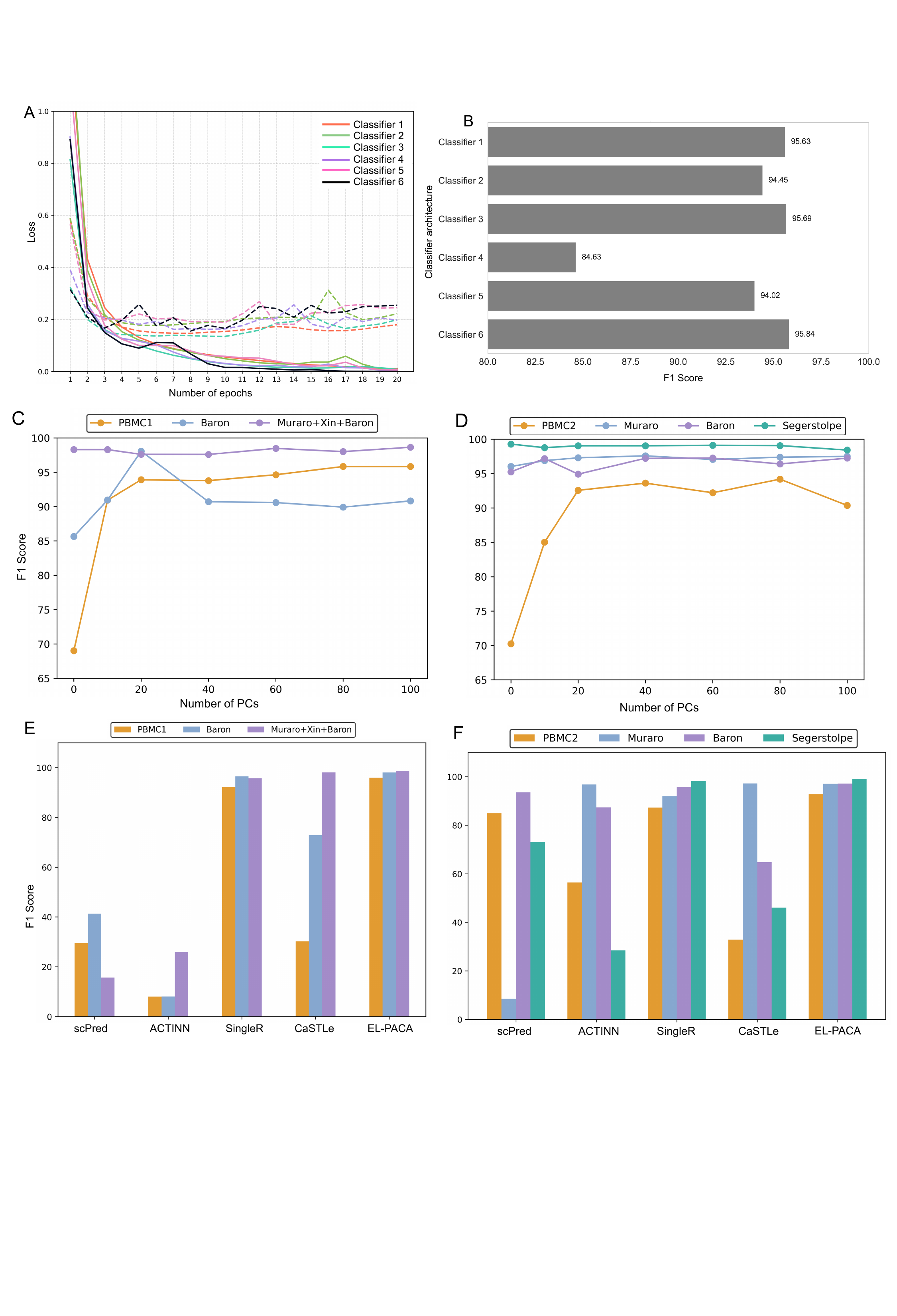}

\subsection*{Figure 2. EL-PACA optimal classifier, robustness and performance across ground truth datasets.}
\textbf{A)} Training and validation loss curves for six different classifier architectures (Classifiers 1–6, each with varying numbers of layers and neurons; see Table S1 for details). Solid lines indicate training loss (error measured on the training data), while dashed lines show validation loss (error measured on a held-out dataset to assess model generalization). The x-axis displays the number of training epochs, and the y-axis represents the loss value for both training and validation. All curves are generated using the PBMC1 dataset. \textbf{B)} Bar plot showing F1 scores (x-axis) as a metric of predictions in the held-out dataset for each classifier architecture (y-axis; detailed in Table S1). \textbf{C)} Graph representing F1 scores as a function of the number of principal components (PCs) for reference-to-reference prediction, where both training and testing data are drawn from the same dataset. The x-axis shows the number of PCs (combined with MDA components), and the y-axis displays the corresponding F1 score across datasets. \textbf{D)} Graph displaying F1 scores for reference-to-query prediction, illustrating how varying the number of PCs influences performance when predicting cell types in an independent query dataset. The x-axis indicates the number of PCs, while the y-axis shows the resulting F1 score. \textbf{E)} Bar charts demonstrating the performance of EL-PACA and current state-of-the-art methods for reference-to-reference prediction, measured by the F1 score. \textbf{F)} Bar charts indicate the performance metrics in predicting query profiles from the reference dataset after applying different existing approaches and EL-PACA using unseen data (reference-to-query) (Figure 1B).

\newpage
\noindent\includegraphics[width=1\linewidth]{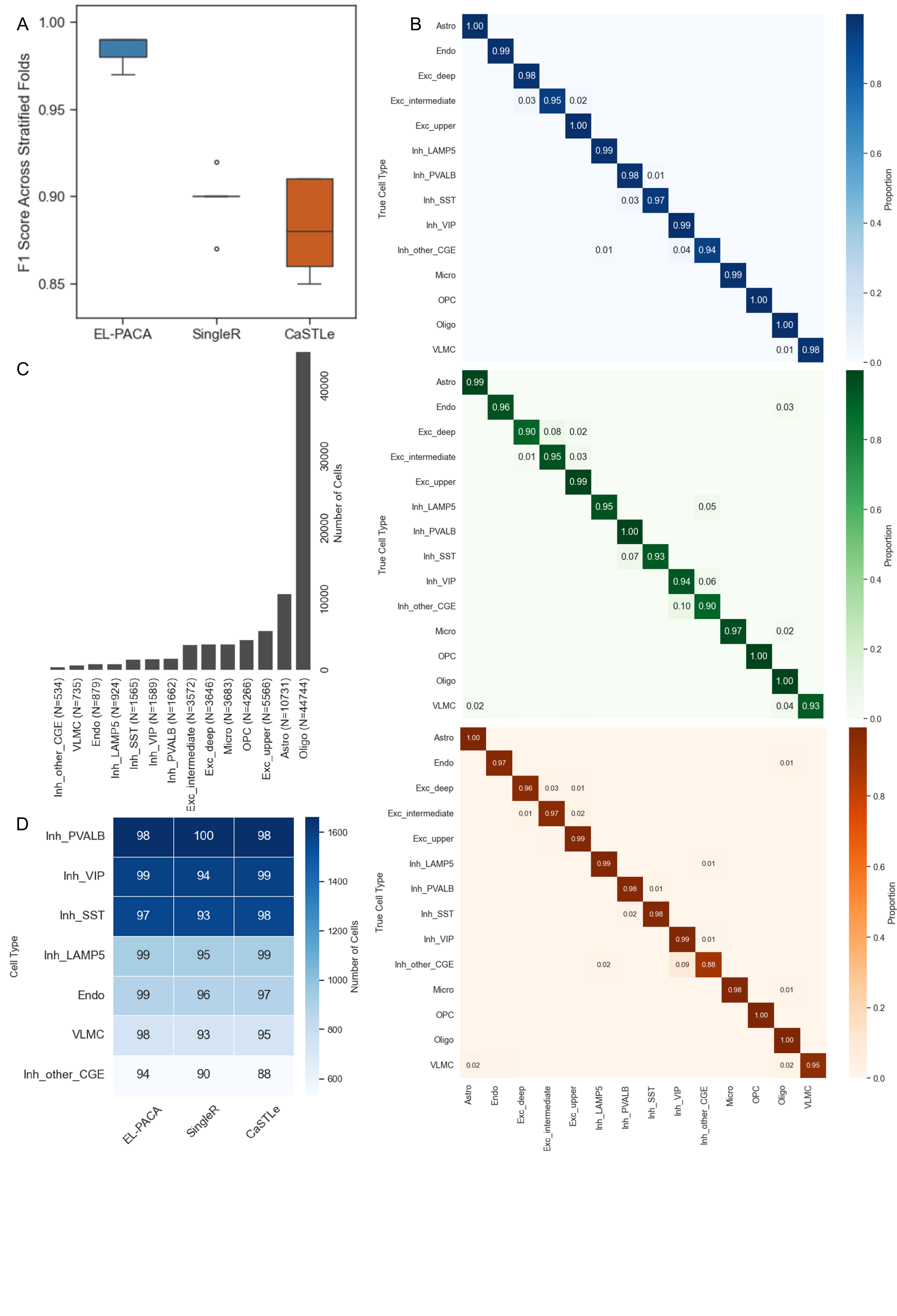}

\subsection*{Figure 3. Comparative performance of cell type annotation methods on ALS postmortem brain single-cell transcriptomic data.}
\textbf{A} Box plot shows the overall classification performance of each annotation method. Each method was trained and tested using stratified five-fold cross-validation. The spread of the box plot indicates variability across the five folds. \textbf{B} Heatmap showing the performance of each method in predicting cell types. The blue heatmap represents the EL-PACA cell-type–level performance, while the green and orange heatmaps represent the performances of SingleR and CaSTLe, respectively. The diagonal cells indicate correct classification rates, whereas the off-diagonal cells represent misclassifications. \textbf{C}  Bar charts showing the number of cells per cell type. \textbf{D} Heatmap representing the performance of EL-PACA, SingleR, and CaSTLe in predicting rare and underrepresented cell types with a smaller number of cells. The values in the heatmap cells represent the percentage of correctly identified cells for each cell type by the corresponding method. Labels: Astro (Astrocytes), Endo (Endothelial Cells), Exc\_deep (Deep-Layer Excitatory Neurons), Exc\_intermediate (Intermediate Excitatory Neurons), Exc\_upper (Upper-Layer Excitatory Neurons), Inh\_LAMP5 (LAMP5-Expressing Inhibitory Neurons), Inh\_PVALB (Parvalbumin-Expressing Inhibitory Neurons), Inh\_SST (Somatostatin-Expressing Inhibitory Neurons), Inh\_VIP (VIP-Expressing Inhibitory Neurons), Inh\_other\_CGE (Other CGE-Derived Inhibitory Neurons), Micro (Microglia), OPC (Oligodendrocyte Progenitor Cells), Oligo (Oligodendrocytes), and VLMC (Vascular and Leptomeningeal Cells).

\newpage

\noindent\includegraphics[width=1\linewidth]{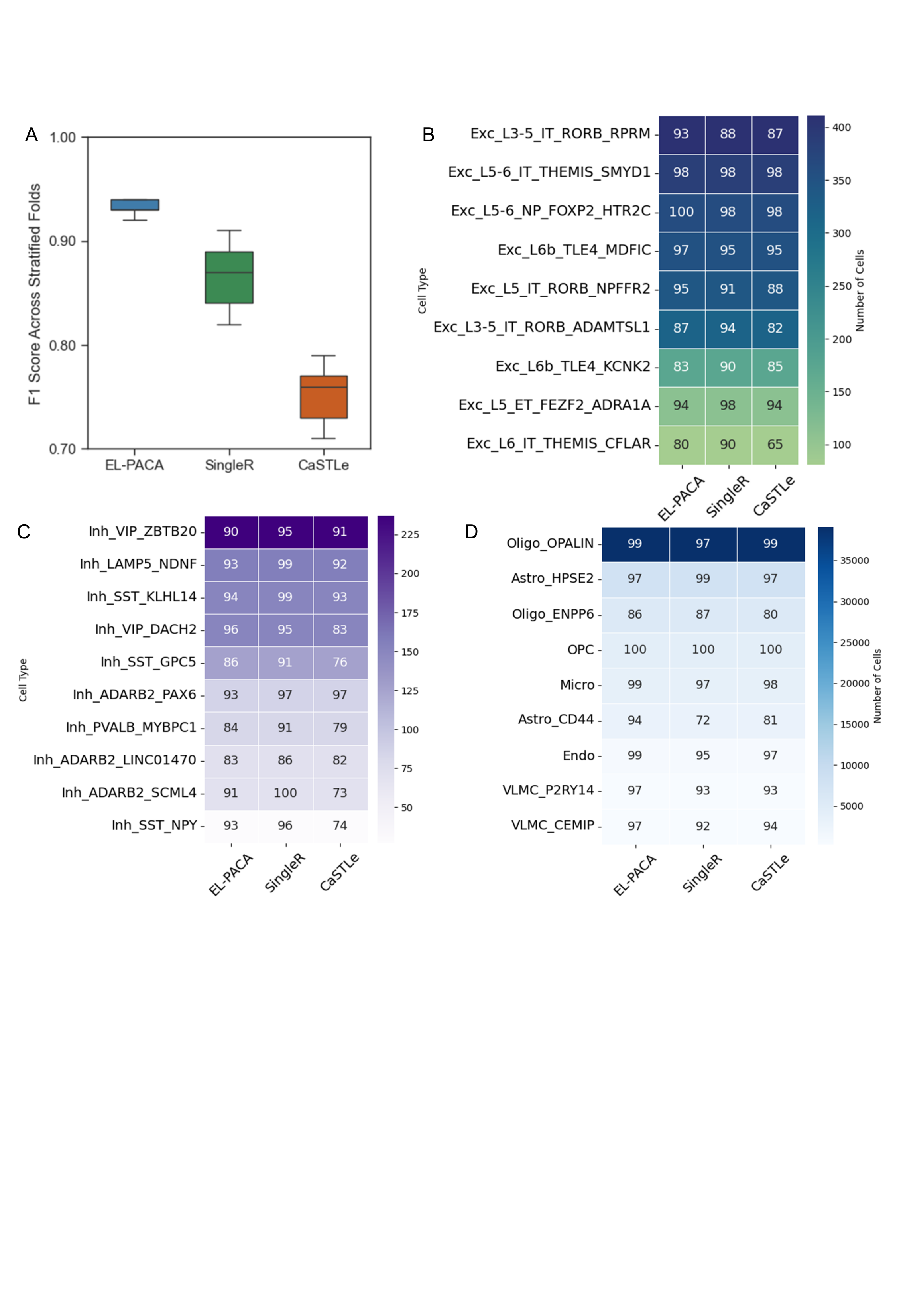}

\subsection*{Figure 4. Comparative performance of cell type annotation methods on low-abundance populations in ALS postmortem samples.}
\textbf{A} Boxplots of classification performance for each method, evaluated using stratified five-fold cross-validation with the F1 score as the metric; boxplot spread reflects variability across folds. \textbf{B} Heatmaps for the ten excitatory neuron subtypes with the fewest cells, comparing EL-PACA, SingleR, and CaSTLe. Tile color intensity encodes the number of cells per subtype; overlaid values indicate the percentage correctly classified. \textbf{C}  Analogous heatmaps for the ten inhibitory neuron subtypes with the fewest cells, with color intensity indicating cell counts and overlaid values showing per-subtype accuracy. \textbf{D} Summary performance for non-neuronal cell types, reported as F1 scores for each method.

\newpage

\noindent\includegraphics[width=1\linewidth]{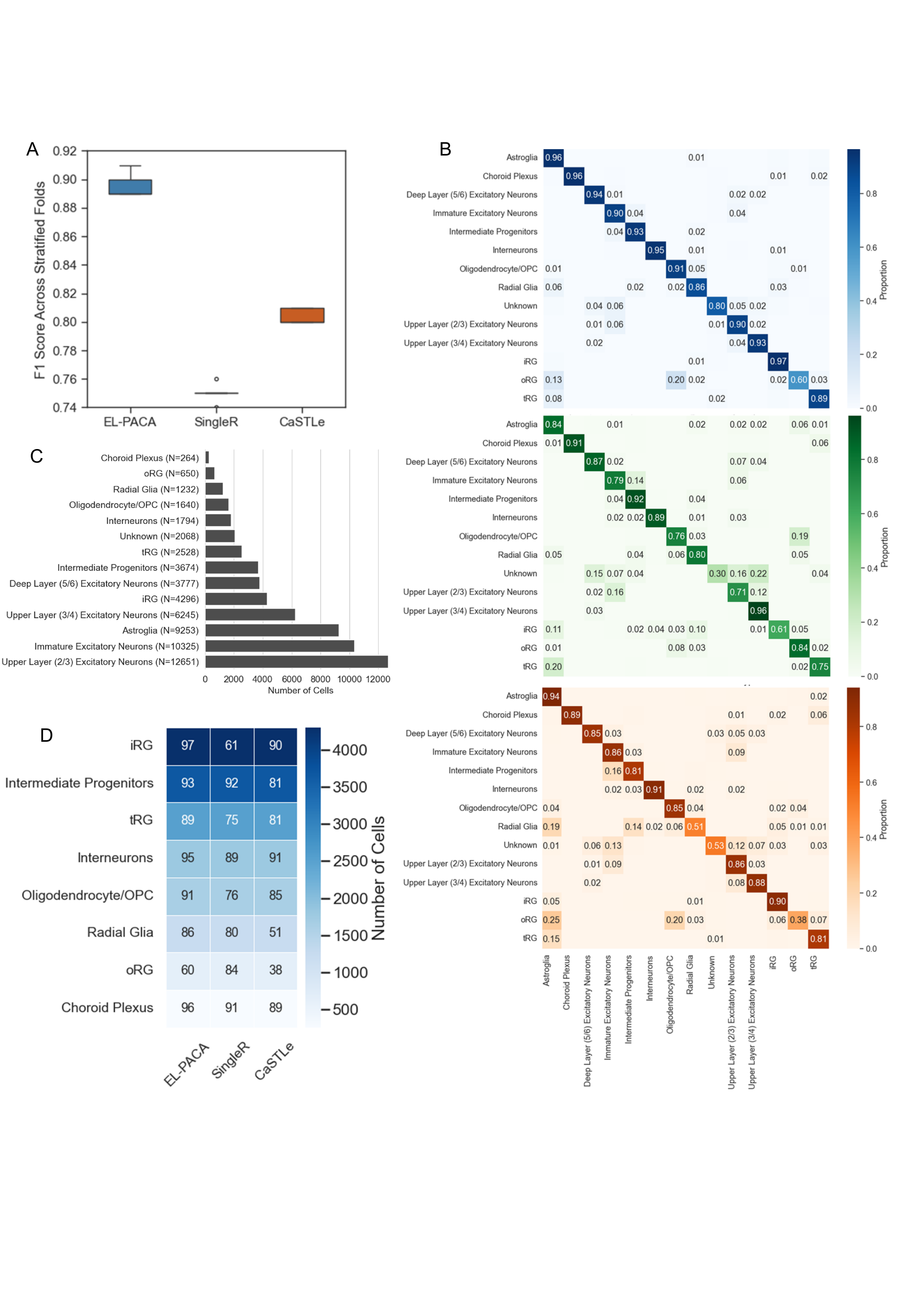}

\subsection*{Figure 5. Comparison of cell type annotation methods on ALS brain organoid transcriptomic data.}
\textbf{A} Overall performance of each annotation method, evaluated using stratified five-fold cross-validation with the F1 score as the metric. The width of the box plots reflects the variability across the five folds. \textbf{B} Cell type-specific performance of EL-PACA, SingleR, and CaSTLe, visualized as blue, green, and orange heatmaps, respectively. Diagonal values represent the proportion of cells accurately classified by each method, while off-diagonal values indicate the extent of misclassification into other cell types. A denser matrix reflects a higher rate of misclassification. \textbf{C}  Total number of cells in each cell type. \textbf{D} Effectiveness of EL-PACA, SingleR, and CaSTLe in detecting rare and underrepresented cell types. Cell types containing a few hundred up to more than 4000 cells were selected to compare each method's performance on rare populations in ALS brain organoids. Each heatmap cell displays the percentage of cells in each category that were correctly classified by the corresponding method.

\newpage

\newpage


\bibliography{references}

\begin{thebibliography}{29}
\providecommand{\natexlab}[1]{#1}
\providecommand{\url}[1]{\texttt{#1}}
\providecommand{\href}[2]{#2}
\providecommand{\path}[1]{#1}
\providecommand{\DOIprefix}{doi: }
\providecommand{\ArXivprefix}{arXiv: }
\providecommand{\URLprefix}{URL: }
\providecommand{\Pubmedprefix}{pmid: }
\providecommand{\doi}[1]{\href{http://dx.doi.org/#1}{\path{#1}}}
\providecommand{\Pubmed}[1]{\href{pmid:#1}{\path{#1}}}
\providecommand{\BIBand}{and}
\providecommand{\bibinfo}[2]{#2}
\ifx\xfnm\undefined \def\xfnm[#1]{\unskip,\space#1}\fi
\makeatletter\def\@biblabel#1{#1.}\makeatother
\bibitem[{Han et~al.(2020)Han, Zhou, Fei, Sun, Wang, Chen, Chen, Wang, Tang, Ge et~al.}]{r1}
\bibinfo{author}{Han, X.}, \bibinfo{author}{Zhou, Z.}, \bibinfo{author}{Fei, L.}, \bibinfo{author}{Sun, H.}, \bibinfo{author}{Wang, R.}, \bibinfo{author}{Chen, Y.}, \bibinfo{author}{Chen, H.}, \bibinfo{author}{Wang, J.}, \bibinfo{author}{Tang, H.}, \bibinfo{author}{Ge, W.} et~al. (\bibinfo{year}{2020}). \bibinfo{title}{Construction of a human cell landscape at single-cell level}.
\newblock \bibinfo{journal}{Nature} \emph{\bibinfo{volume}{581}}, \bibinfo{pages}{303--309}.
\bibitem[{Gulati et~al.(2020)Gulati, Sikandar, Wesche, Manjunath, Bharadwaj, Berger, Ilagan, Kuo, Hsieh, Cai et~al.}]{r3}
\bibinfo{author}{Gulati, G.S.}, \bibinfo{author}{Sikandar, S.S.}, \bibinfo{author}{Wesche, D.J.}, \bibinfo{author}{Manjunath, A.}, \bibinfo{author}{Bharadwaj, A.}, \bibinfo{author}{Berger, M.J.}, \bibinfo{author}{Ilagan, F.}, \bibinfo{author}{Kuo, A.H.}, \bibinfo{author}{Hsieh, R.W.}, \bibinfo{author}{Cai, S.} et~al. (\bibinfo{year}{2020}). \bibinfo{title}{Single-cell transcriptional diversity is a hallmark of developmental potential}.
\newblock \bibinfo{journal}{Science} \emph{\bibinfo{volume}{367}}, \bibinfo{pages}{405--411}.
\bibitem[{Clarence et~al.(2025)Clarence, Bendl, Cao, Wang, Zheng, Hoffman, Kozlenkov, Hong, Iskhakova, Jaiswal et~al.}]{r5}
\bibinfo{author}{Clarence, T.}, \bibinfo{author}{Bendl, J.}, \bibinfo{author}{Cao, X.}, \bibinfo{author}{Wang, X.}, \bibinfo{author}{Zheng, S.}, \bibinfo{author}{Hoffman, G.E.}, \bibinfo{author}{Kozlenkov, A.}, \bibinfo{author}{Hong, A.}, \bibinfo{author}{Iskhakova, M.}, \bibinfo{author}{Jaiswal, M.K.} et~al. (\bibinfo{year}{2025}). \bibinfo{title}{Multiomic single-cell profiling identifies critical regulators of postnatal brain}.
\newblock \bibinfo{journal}{Nature Genetics} pp. \bibinfo{pages}{1--13}.
\bibitem[{Pineda et~al.(2024)Pineda, Lee, Ulloa-Navas, Linville, Garcia, Galani, Engelberg-Cook, Castanedes, Fitzwalter, Pregent et~al.}]{r6}
\bibinfo{author}{Pineda, S.S.}, \bibinfo{author}{Lee, H.}, \bibinfo{author}{Ulloa-Navas, M.J.}, \bibinfo{author}{Linville, R.M.}, \bibinfo{author}{Garcia, F.J.}, \bibinfo{author}{Galani, K.}, \bibinfo{author}{Engelberg-Cook, E.}, \bibinfo{author}{Castanedes, M.C.}, \bibinfo{author}{Fitzwalter, B.E.}, \bibinfo{author}{Pregent, L.J.} et~al. (\bibinfo{year}{2024}). \bibinfo{title}{Single-cell dissection of the human motor and prefrontal cortices in als and ftld}.
\newblock \bibinfo{journal}{Cell} \emph{\bibinfo{volume}{187}}, \bibinfo{pages}{1971--1989}.
\bibitem[{Li et~al.(2023)Li, Jaiswal, Chien, Kozlenkov, Jung, Zhou, Gardashli, Pregent, Engelberg-Cook, Dickson et~al.}]{r7}
\bibinfo{author}{Li, J.}, \bibinfo{author}{Jaiswal, M.K.}, \bibinfo{author}{Chien, J.F.}, \bibinfo{author}{Kozlenkov, A.}, \bibinfo{author}{Jung, J.}, \bibinfo{author}{Zhou, P.}, \bibinfo{author}{Gardashli, M.}, \bibinfo{author}{Pregent, L.J.}, \bibinfo{author}{Engelberg-Cook, E.}, \bibinfo{author}{Dickson, D.W.} et~al. (\bibinfo{year}{2023}). \bibinfo{title}{Divergent single cell transcriptome and epigenome alterations in als and ftd patients with c9orf72 mutation}.
\newblock \bibinfo{journal}{Nature communications} \emph{\bibinfo{volume}{14}}, \bibinfo{pages}{5714}.
\bibitem[{Zhang et~al.(2019)Zhang, Lan, Xu, Quan, Zhao, Deng, Luo, Xu, Liao, Yan et~al.}]{r8}
\bibinfo{author}{Zhang, X.}, \bibinfo{author}{Lan, Y.}, \bibinfo{author}{Xu, J.}, \bibinfo{author}{Quan, F.}, \bibinfo{author}{Zhao, E.}, \bibinfo{author}{Deng, C.}, \bibinfo{author}{Luo, T.}, \bibinfo{author}{Xu, L.}, \bibinfo{author}{Liao, G.}, \bibinfo{author}{Yan, M.} et~al. (\bibinfo{year}{2019}). \bibinfo{title}{Cellmarker: a manually curated resource of cell markers in human and mouse}.
\newblock \bibinfo{journal}{Nucleic acids research} \emph{\bibinfo{volume}{47}}, \bibinfo{pages}{D721--D728}.
\bibitem[{Aran et~al.(2019)Aran, Looney, Liu, Wu, Fong, Hsu, Chak, Naikawadi, Wolters, Abate et~al.}]{r9}
\bibinfo{author}{Aran, D.}, \bibinfo{author}{Looney, A.P.}, \bibinfo{author}{Liu, L.}, \bibinfo{author}{Wu, E.}, \bibinfo{author}{Fong, V.}, \bibinfo{author}{Hsu, A.}, \bibinfo{author}{Chak, S.}, \bibinfo{author}{Naikawadi, R.P.}, \bibinfo{author}{Wolters, P.J.}, \bibinfo{author}{Abate, A.R.} et~al. (\bibinfo{year}{2019}). \bibinfo{title}{Reference-based analysis of lung single-cell sequencing reveals a transitional profibrotic macrophage}.
\newblock \bibinfo{journal}{Nature immunology} \emph{\bibinfo{volume}{20}}, \bibinfo{pages}{163--172}.
\bibitem[{Alquicira-Hernandez et~al.(2019)Alquicira-Hernandez, Sathe, Ji, Nguyen and Powell}]{r10}
\bibinfo{author}{Alquicira-Hernandez, J.}, \bibinfo{author}{Sathe, A.}, \bibinfo{author}{Ji, H.P.}, \bibinfo{author}{Nguyen, Q.}, and \bibinfo{author}{Powell, J.E.} (\bibinfo{year}{2019}). \bibinfo{title}{scpred: accurate supervised method for cell-type classification from single-cell rna-seq data}.
\newblock \bibinfo{journal}{Genome biology} \emph{\bibinfo{volume}{20}}, \bibinfo{pages}{264}.
\bibitem[{Lieberman et~al.(2018)Lieberman, Rokach and Shay}]{r11}
\bibinfo{author}{Lieberman, Y.}, \bibinfo{author}{Rokach, L.}, and \bibinfo{author}{Shay, T.} (\bibinfo{year}{2018}). \bibinfo{title}{Castle--classification of single cells by transfer learning: harnessing the power of publicly available single cell rna sequencing experiments to annotate new experiments}.
\newblock \bibinfo{journal}{PloS one} \emph{\bibinfo{volume}{13}}, \bibinfo{pages}{e0205499}.
\bibitem[{Ma and Pellegrini(2020)}]{r12}
\bibinfo{author}{Ma, F.}, and \bibinfo{author}{Pellegrini, M.} (\bibinfo{year}{2020}). \bibinfo{title}{Actinn: automated identification of cell types in single cell rna sequencing}.
\newblock \bibinfo{journal}{Bioinformatics} \emph{\bibinfo{volume}{36}}, \bibinfo{pages}{533--538}.
\bibitem[{Zeng et~al.(2025)Zeng, Xie, Shangguan, Wei, Li, Su, Yang, Zhang, Zhang, Fang et~al.}]{r13}
\bibinfo{author}{Zeng, Y.}, \bibinfo{author}{Xie, J.}, \bibinfo{author}{Shangguan, N.}, \bibinfo{author}{Wei, Z.}, \bibinfo{author}{Li, W.}, \bibinfo{author}{Su, Y.}, \bibinfo{author}{Yang, S.}, \bibinfo{author}{Zhang, C.}, \bibinfo{author}{Zhang, J.}, \bibinfo{author}{Fang, N.} et~al. (\bibinfo{year}{2025}). \bibinfo{title}{Cellfm: a large-scale foundation model pre-trained on transcriptomics of 100 million human cells}.
\newblock \bibinfo{journal}{Nature Communications} \emph{\bibinfo{volume}{16}}, \bibinfo{pages}{4679}.
\bibitem[{Hao et~al.(2024{\natexlab{a}})Hao, Gong, Zeng, Liu, Guo, Cheng, Wang, Ma, Zhang and Song}]{r15}
\bibinfo{author}{Hao, M.}, \bibinfo{author}{Gong, J.}, \bibinfo{author}{Zeng, X.}, \bibinfo{author}{Liu, C.}, \bibinfo{author}{Guo, Y.}, \bibinfo{author}{Cheng, X.}, \bibinfo{author}{Wang, T.}, \bibinfo{author}{Ma, J.}, \bibinfo{author}{Zhang, X.}, and \bibinfo{author}{Song, L.} (\bibinfo{year}{2024}{\natexlab{a}}). \bibinfo{title}{Large-scale foundation model on single-cell transcriptomics}.
\newblock \bibinfo{journal}{Nature methods} \emph{\bibinfo{volume}{21}}, \bibinfo{pages}{1481--1491}.
\bibitem[{McKeever et~al.(2023)McKeever, Sababi, Sharma, Khuu, Xu, Shen, Xiao, McGoldrick, Orouji, Ketela et~al.}]{r23}
\bibinfo{author}{McKeever, P.M.}, \bibinfo{author}{Sababi, A.M.}, \bibinfo{author}{Sharma, R.}, \bibinfo{author}{Khuu, N.}, \bibinfo{author}{Xu, Z.}, \bibinfo{author}{Shen, S.Y.}, \bibinfo{author}{Xiao, S.}, \bibinfo{author}{McGoldrick, P.}, \bibinfo{author}{Orouji, E.}, \bibinfo{author}{Ketela, T.} et~al. (\bibinfo{year}{2023}). \bibinfo{title}{Single-nucleus multiomic atlas of frontal cortex in amyotrophic lateral sclerosis with a deep learning-based decoding of alternative polyadenylation mechanisms}.
\newblock \bibinfo{journal}{bioRxiv} pp. \bibinfo{pages}{2023--12}.
\bibitem[{Kamath et~al.(2022)Kamath, Abdulraouf, Burris, Langlieb, Gazestani, Nadaf, Balderrama, Vanderburg and Macosko}]{r24}
\bibinfo{author}{Kamath, T.}, \bibinfo{author}{Abdulraouf, A.}, \bibinfo{author}{Burris, S.}, \bibinfo{author}{Langlieb, J.}, \bibinfo{author}{Gazestani, V.}, \bibinfo{author}{Nadaf, N.M.}, \bibinfo{author}{Balderrama, K.}, \bibinfo{author}{Vanderburg, C.}, and \bibinfo{author}{Macosko, E.Z.} (\bibinfo{year}{2022}). \bibinfo{title}{Single-cell genomic profiling of human dopamine neurons identifies a population that selectively degenerates in parkinson’s disease}.
\newblock \bibinfo{journal}{Nature neuroscience} \emph{\bibinfo{volume}{25}}, \bibinfo{pages}{588--595}.
\bibitem[{Rexach et~al.(2024)Rexach, Cheng, Chen, Polioudakis, Lin, Mitri, Elkins, Han, Yamakawa, Yin et~al.}]{r25}
\bibinfo{author}{Rexach, J.E.}, \bibinfo{author}{Cheng, Y.}, \bibinfo{author}{Chen, L.}, \bibinfo{author}{Polioudakis, D.}, \bibinfo{author}{Lin, L.C.}, \bibinfo{author}{Mitri, V.}, \bibinfo{author}{Elkins, A.}, \bibinfo{author}{Han, X.}, \bibinfo{author}{Yamakawa, M.}, \bibinfo{author}{Yin, A.} et~al. (\bibinfo{year}{2024}). \bibinfo{title}{Cross-disorder and disease-specific pathways in dementia revealed by single-cell genomics}.
\newblock \bibinfo{journal}{Cell} \emph{\bibinfo{volume}{187}}, \bibinfo{pages}{5753--5774}.
\bibitem[{Szeb{\'e}nyi et~al.(2021)Szeb{\'e}nyi, Wenger, Sun, Dunn, Limegrover, Gibbons, Conci, Paulsen, Mierau, Balmus et~al.}]{r16}
\bibinfo{author}{Szeb{\'e}nyi, K.}, \bibinfo{author}{Wenger, L.M.}, \bibinfo{author}{Sun, Y.}, \bibinfo{author}{Dunn, A.W.}, \bibinfo{author}{Limegrover, C.A.}, \bibinfo{author}{Gibbons, G.M.}, \bibinfo{author}{Conci, E.}, \bibinfo{author}{Paulsen, O.}, \bibinfo{author}{Mierau, S.B.}, \bibinfo{author}{Balmus, G.} et~al. (\bibinfo{year}{2021}). \bibinfo{title}{Human als/ftd brain organoid slice cultures display distinct early astrocyte and targetable neuronal pathology}.
\newblock \bibinfo{journal}{Nature neuroscience} \emph{\bibinfo{volume}{24}}, \bibinfo{pages}{1542--1554}.
\bibitem[{Azbukina et~al.(2025)Azbukina, He, Lin, Santel, Kashanian, Maynard, T{\"o}r{\"o}k, Okamoto, Nikolova, Kanton et~al.}]{r26}
\bibinfo{author}{Azbukina, N.}, \bibinfo{author}{He, Z.}, \bibinfo{author}{Lin, H.C.}, \bibinfo{author}{Santel, M.}, \bibinfo{author}{Kashanian, B.}, \bibinfo{author}{Maynard, A.}, \bibinfo{author}{T{\"o}r{\"o}k, T.}, \bibinfo{author}{Okamoto, R.}, \bibinfo{author}{Nikolova, M.}, \bibinfo{author}{Kanton, S.} et~al. (\bibinfo{year}{2025}). \bibinfo{title}{Multi-omic human neural organoid cell atlas of the posterior brain}.
\newblock \bibinfo{journal}{bioRxiv} pp. \bibinfo{pages}{2025--03}.
\bibitem[{Korsunsky et~al.(2019)Korsunsky, Millard, Fan, Slowikowski, Zhang, Wei, Baglaenko, Brenner, Loh and Raychaudhuri}]{r31}
\bibinfo{author}{Korsunsky, I.}, \bibinfo{author}{Millard, N.}, \bibinfo{author}{Fan, J.}, \bibinfo{author}{Slowikowski, K.}, \bibinfo{author}{Zhang, F.}, \bibinfo{author}{Wei, K.}, \bibinfo{author}{Baglaenko, Y.}, \bibinfo{author}{Brenner, M.}, \bibinfo{author}{Loh, P.r.}, and \bibinfo{author}{Raychaudhuri, S.} (\bibinfo{year}{2019}). \bibinfo{title}{Fast, sensitive and accurate integration of single-cell data with harmony}.
\newblock \bibinfo{journal}{Nature methods} \emph{\bibinfo{volume}{16}}, \bibinfo{pages}{1289--1296}.
\bibitem[{Hao et~al.(2024{\natexlab{b}})Hao, Stuart, Kowalski, Choudhary, Hoffman, Hartman, Srivastava, Molla, Madad, Fernandez-Granda et~al.}]{r17}
\bibinfo{author}{Hao, Y.}, \bibinfo{author}{Stuart, T.}, \bibinfo{author}{Kowalski, M.H.}, \bibinfo{author}{Choudhary, S.}, \bibinfo{author}{Hoffman, P.}, \bibinfo{author}{Hartman, A.}, \bibinfo{author}{Srivastava, A.}, \bibinfo{author}{Molla, G.}, \bibinfo{author}{Madad, S.}, \bibinfo{author}{Fernandez-Granda, C.} et~al. (\bibinfo{year}{2024}{\natexlab{b}}). \bibinfo{title}{Dictionary learning for integrative, multimodal and scalable single-cell analysis}.
\newblock \bibinfo{journal}{Nature biotechnology} \emph{\bibinfo{volume}{42}}, \bibinfo{pages}{293--304}.
\bibitem[{Wolf et~al.(2018)Wolf, Angerer and Theis}]{r18}
\bibinfo{author}{Wolf, F.A.}, \bibinfo{author}{Angerer, P.}, and \bibinfo{author}{Theis, F.J.} (\bibinfo{year}{2018}). \bibinfo{title}{Scanpy: large-scale single-cell gene expression data analysis}.
\newblock \bibinfo{journal}{Genome biology} \emph{\bibinfo{volume}{19}}, \bibinfo{pages}{15}.
\bibitem[{Abdelaal et~al.(2019)Abdelaal, Michielsen, Cats, Hoogduin, Mei, Reinders and Mahfouz}]{r27}
\bibinfo{author}{Abdelaal, T.}, \bibinfo{author}{Michielsen, L.}, \bibinfo{author}{Cats, D.}, \bibinfo{author}{Hoogduin, D.}, \bibinfo{author}{Mei, H.}, \bibinfo{author}{Reinders, M.J.}, and \bibinfo{author}{Mahfouz, A.} (\bibinfo{year}{2019}). \bibinfo{title}{A comparison of automatic cell identification methods for single-cell rna sequencing data}.
\newblock \bibinfo{journal}{Genome biology} \emph{\bibinfo{volume}{20}}, \bibinfo{pages}{194}.
\bibitem[{Dann et~al.(2023)Dann, Cujba, Oliver et~al.}]{r32}
\bibinfo{author}{Dann, E.}, \bibinfo{author}{Cujba, A.M.}, \bibinfo{author}{Oliver, A.J.} et~al. (\bibinfo{year}{2023}). \bibinfo{title}{Precise identification of cell states altered in disease using healthy single-cell references}.
\newblock \bibinfo{journal}{Nature Genetics} \emph{\bibinfo{volume}{55}}, \bibinfo{pages}{1998--2008}. \DOIprefix\doi{10.1038/s41588-023-01523-7}.
\bibitem[{Mathys et~al.(2024)Mathys, Boix, Akay, Xia, Davila-Velderrain, Ng, Jiang, Abdelhady, Galani, Mantero et~al.}]{r28}
\bibinfo{author}{Mathys, H.}, \bibinfo{author}{Boix, C.A.}, \bibinfo{author}{Akay, L.A.}, \bibinfo{author}{Xia, Z.}, \bibinfo{author}{Davila-Velderrain, J.}, \bibinfo{author}{Ng, A.P.}, \bibinfo{author}{Jiang, X.}, \bibinfo{author}{Abdelhady, G.}, \bibinfo{author}{Galani, K.}, \bibinfo{author}{Mantero, J.} et~al. (\bibinfo{year}{2024}). \bibinfo{title}{Single-cell multiregion dissection of alzheimer’s disease}.
\newblock \bibinfo{journal}{Nature} \emph{\bibinfo{volume}{632}}, \bibinfo{pages}{858--868}.
\bibitem[{Park et~al.(2023)Park, Cho, Kim, Saito, Saido, Won and Kim}]{r29}
\bibinfo{author}{Park, H.}, \bibinfo{author}{Cho, B.}, \bibinfo{author}{Kim, H.}, \bibinfo{author}{Saito, T.}, \bibinfo{author}{Saido, T.C.}, \bibinfo{author}{Won, K.J.}, and \bibinfo{author}{Kim, J.} (\bibinfo{year}{2023}). \bibinfo{title}{Single-cell rna-sequencing identifies disease-associated oligodendrocytes in male app nl-gf and 5xfad mice}.
\newblock \bibinfo{journal}{Nature Communications} \emph{\bibinfo{volume}{14}}, \bibinfo{pages}{802}.
\bibitem[{He et~al.(2024)He, Dony, Fleck et~al.}]{r33}
\bibinfo{author}{He, Z.}, \bibinfo{author}{Dony, L.}, \bibinfo{author}{Fleck, J.S.} et~al. (\bibinfo{year}{2024}). \bibinfo{title}{An integrated transcriptomic cell atlas of human neural organoids}.
\newblock \bibinfo{journal}{Nature} \emph{\bibinfo{volume}{635}}, \bibinfo{pages}{690--698}. \DOIprefix\doi{10.1038/s41586-024-08172-8}.
\bibitem[{Caporale et~al.(2025)Caporale, Castaldi, Rigoli et~al.}]{r34}
\bibinfo{author}{Caporale, N.}, \bibinfo{author}{Castaldi, D.}, \bibinfo{author}{Rigoli, M.T.} et~al. (\bibinfo{year}{2025}). \bibinfo{title}{Multiplexing cortical brain organoids for the longitudinal dissection of developmental traits at single-cell resolution}.
\newblock \bibinfo{journal}{Nature Methods} \emph{\bibinfo{volume}{22}}, \bibinfo{pages}{358--370}. \DOIprefix\doi{10.1038/s41592-024-02555-5}.
\bibitem[{Fleck et~al.(2023)Fleck, Jansen, Wollny et~al.}]{r35}
\bibinfo{author}{Fleck, J.S.}, \bibinfo{author}{Jansen, S.M.J.}, \bibinfo{author}{Wollny, D.} et~al. (\bibinfo{year}{2023}). \bibinfo{title}{Inferring and perturbing cell fate regulomes in human brain organoids}.
\newblock \bibinfo{journal}{Nature} \emph{\bibinfo{volume}{621}}, \bibinfo{pages}{365--372}. \DOIprefix\doi{10.1038/s41586-022-05279-8}.
\bibitem[{Marton et~al.(2019)Marton, Miura, Sloan, Li, Revah, Levy, Huguenard and Pa{\c{s}}ca}]{r36}
\bibinfo{author}{Marton, R.M.}, \bibinfo{author}{Miura, Y.}, \bibinfo{author}{Sloan, S.A.}, \bibinfo{author}{Li, Q.}, \bibinfo{author}{Revah, O.}, \bibinfo{author}{Levy, R.J.}, \bibinfo{author}{Huguenard, J.R.}, and \bibinfo{author}{Pa{\c{s}}ca, S.P.} (\bibinfo{year}{2019}). \bibinfo{title}{Differentiation and maturation of oligodendrocytes in human three-dimensional neural cultures}.
\newblock \bibinfo{journal}{Nature Neuroscience} \emph{\bibinfo{volume}{22}}, \bibinfo{pages}{484--491}. \DOIprefix\doi{10.1038/s41593-018-0316-9}.
\bibitem[{Rosen et~al.(2023)Rosen, Roohani, Agarwal, Samotor{\v{c}}an, Consortium, Quake and Leskovec}]{r14}
\bibinfo{author}{Rosen, Y.}, \bibinfo{author}{Roohani, Y.}, \bibinfo{author}{Agarwal, A.}, \bibinfo{author}{Samotor{\v{c}}an, L.}, \bibinfo{author}{Consortium, T.S.}, \bibinfo{author}{Quake, S.R.}, and \bibinfo{author}{Leskovec, J.} (\bibinfo{year}{2023}). \bibinfo{title}{Universal cell embeddings: A foundation model for cell biology}.
\newblock \bibinfo{journal}{bioRxiv} pp. \bibinfo{pages}{2023--11}.

\end{thebibliography}

\bigskip


\newpage

\end{document}